\documentclass{article}

\PassOptionsToPackage{numbers, compress}{natbib}

\usepackage{wrapfig}

\usepackage[final]{main}

\usepackage[utf8]{inputenc} 
\usepackage[T1]{fontenc}    
\usepackage{hyperref}       
\usepackage{url}            
\usepackage{booktabs}       
\usepackage{amsfonts}       
\usepackage{nicefrac}       
\usepackage{microtype}      
\usepackage{tikz} 
\usetikzlibrary{shapes.geometric, arrows}

 \usepackage[%
  capitalize,
  sort&compress
]{cleveref}

\usepackage{ulem}

\title{An Enriched Automated PV Registry: Combining Image Recognition and 3D Building Data}

\author{%
  Benjamin Rausch*$^{1,2}$, Kevin Mayer*$^1$, Marie-Louise Arlt$^{1,3}$, Gunther Gust$^3$ \\\textbf{Philipp Staudt$^2$, Christof Weinhardt$^2$, Dirk Neumann$^3$, Ram Rajagopal$^1$} \\
  Stanford University$^1$, Karlsruhe Institute of Technology$^2$, University of Freiburg$^3$\\
  \texttt{\{rauschbe, kdmayer, mlarlt, ramr\}@stanford.edu} \\
  \texttt{\{philipp.staudt, christof.weinhardt\}@kit.edu}\\
    \texttt{\{gunther.gust, dirk.neumann\}@is.uni-freiburg.de}
}

\begin{document}

\maketitle

\begin{abstract}

While photovoltaic~(PV) systems are installed at an unprecedented rate, reliable information on an installation level remains scarce. As a result, automatically created PV registries are a timely contribution to optimize grid planning and operations. This paper demonstrates how aerial imagery and three-dimensional building data can be combined to create an address-level PV registry, specifying area, tilt, and orientation angles. We demonstrate the benefits of this approach for PV capacity estimation. In addition, this work presents, for the first time, a comparison between automated and officially-created PV registries. Our results indicate that our enriched automated registry proves to be useful to validate, update, and complement official registries.
\end{abstract}

\footnotetext{*Equal contribution.}
\section{Introduction and Related Work}

Photovoltaics (PV) is a key technology to decarbonize our energy systems. Hence, knowing the location, size, and orientation of existing PV systems is essential to predict PV electricity supply and to support the planning and operation of electricity systems. However, keeping track of installed PV systems is difficult due to their decentralized character and large number.

Centralized PV registries are intended to keep track of PV system installations but have considerable shortcomings. One approach to collect PV system data are self-reports, such as the Open PV project \cite{openpv} or Germany's official PV registry \cite{mastr}, where owners are required to register their PV system. Efforts for generating and maintaining these registries, however, are high and human data entry is often error-prone. Furthermore, PV systems are mostly registered using street addresses so that important information on the exact location of the systems, such as corresponding rooftops, remain unknown. 
To address these shortcomings, a second type of registries has recently emerged, which recollects PV system data in an automated fashion using aerial imagery and machine learning \citep{cnnsolar1,Yu2018,CBS2020,Mayer2020}. An overview is provided in \citep{deHoog2020}. Yet, these approaches do not link detected PV systems to their respective buildings. Hence, important information for power systems operation, such as tilt angle, orientation angle, occupied rooftop area, as well as socio-economic information associated with buildings and their owners remains unknown. 

In this work, we first propose an automated approach to generate an enriched PV registry with additional information such as tilt angle, orientation, and general building data. For this purpose, we combine aerial imagery with 3D building geometries. This enriched registry provides the basis for several important applications, such as supply predictions, PV nowcasting, and estimation of available rooftop capacities \cite{deHoog2020}. As a second contribution, we demonstrate how the tilt angle derived from 3D building data allows for an improved PV generation capacity estimation. As a third contribution, we compare our constructed registry to the German PV registry \cite{mastr}---an official registry relying on self-reported data. This has previously not been done \cite{deHoog2020}. We thereby identify non-reported installations, erroneous locations, and duplicated entries. This shows that automated PV registries, such as our approach, may be valuable for policy makers for improving official PV registries relying on self-reported data. 

\section{Methodology}
\begin{figure}[htb]
    \centering
    \includegraphics[width =\columnwidth]{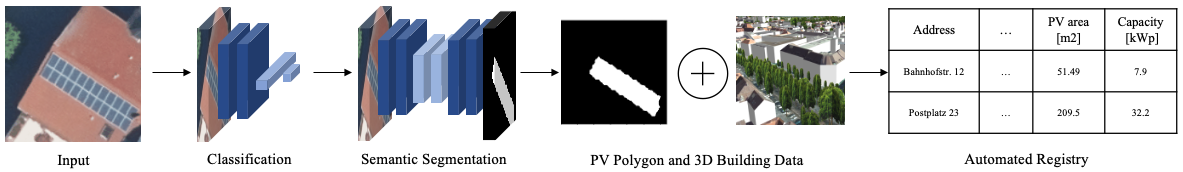}
    \caption{Pipeline to create an automated PV registry.}
    \label{fig:pipeline}
\end{figure}

In \Cref{fig:pipeline}, we describe how aerial imagery and 3D building data are combined in order to create the enriched PV registry with an automated processing pipeline. The pipeline consists of the following steps. 


\textbf{Input.} The processing pipeline uses aerial imagery and 3D building data as input. 
\newline
\newline
\textbf{Classification.} The classification module detects whether or not a PV system is present in an image. For classification, we use a fine-tuned model for Germany which is based on the DeepSolar architecture \cite{Yu2018}, 
as described in Section \ref{sec: traindeepsolar}. Only aerial images classified as depicting PV systems are propagated to the segmentation module. 
\newline
\newline
\textbf{Semantic Segmentation.} The segmentation module determines which areas are covered by a PV system. We use the pre-trained DeepLabv3 CNN with a Resnet-101 backbone for semantic segmentation \cite{chen2017rethinking}. The model is pre-trained on a subset of COCO Train 2017, i.e. the 20 categories that are present in the Pascal VOC dataset. We fine-tune DeepLabv3 for the task of PV system segmentation from aerial imagery. The hyperparameters are set to the best values described in the original work \cite{chen2017rethinking}. Training specifics and related information are described in Section \ref{sec: traindeeplab}. 
\newline
\newline
\textbf{PV Polygon and 3D Building Data.} 
Rolling out the classification and segmentation modules on a large scale, we obtain an instance-level PV system list containing the estimated shapes as real world coordinate polygons. 
After intersecting the PV system polygons with the rooftop polygons, we obtain PV polygons accurately mapped to the corresponding rooftops. Based on this result, we calculate the area occupied by a PV system in $m^2$ and adjust this value by taking the tilt angle into account. 
To compare the results with the official registry, we estimate the installation capacity by assuming a factor of $6.5m^2$ per $kWh$ generation peak \cite{sizepv} with respect to our area estimates.
\newline
\newline
\textbf{Automated Registry.}
Intersecting the PV system area estimates with rooftop polygons enables us to create an automated registry. For each address associated with the rooftop data, we group detected PV systems and aggregate the respective PV panel sizes and capacity estimates. We exclude entries with a segmented PV installation area lower than 6$m^2$ as these indicate processing errors, e.g. minor overlaps of segmented PV installations with adjacent rooftops.


\section{Data}
\label{data}

\textbf{Image Data.}
We gather high resolution aerial imagery with a native spatial resolution of around 0.1 meter per pixel provided by \cite{aerialimagerynrw}. Each image contains $320 \times 320$ pixels and is downloaded at an upsampled spatial resolution of 0.05 meter per pixel. 
The \textit{classification dataset} contains 107,809 images in total. Table \ref{tab:classification_dataset} in the appendix provides an overview of the dataset. While the validation and test set splits are drawn from North Rhine-Westphalia at random, the training set is composed according to the dataset creation strategy described in \cite{Mayer2020} and further enriched with difficult-to-classify samples. 
The \textit{segmentation dataset} consists of 4,028 images in total, which are randomly drawn from the positively classified locations in North Rhine-Westphalia provided by \cite{Mayer2020}. The dataset is further enriched with samples depicting shadows, reflections, and other difficult-to-segment effects. The dataset split is described in Table \ref{tab:segmentation_dataset} in the appendix. 
\newline
\newline
\textbf{3D Building Data.} 
The 3D building data for North Rhine-Westphalia is downloaded from \cite{3ddaten}. It has been created by fusing LiDAR-based point clouds with other publicly available datasets \cite{citygml}, e.g. building footprints. With a height estimation error of less than one meter, it provides highly accurate information about each geometric feature of a building, e.g. wall, rooftop, and footprint polygons. Our work focuses on rooftop polygons in order to extract a rooftop's tilt and orientation referring to \cite{tiltazimuth} for calculation. Finally, each rooftop polygon is flattened from 3D to 2D. 
\newline
\newline
\textbf{Official PV Registry.} 
The German government provides an official registry of generation units listing the address, generation capacity, and other related information (\textit{`Marktstammdatenregister'}). The public version only publishes accurate address information for PV installations larger than 30$kWp$. 
\section{Results}

\subsection{Classification and Segmentation Performance}

In \Cref{tbl: classificationpaper} and  \Cref{tbl: segpaper}, we compare our pipeline's classification and segmentation performance to the state of the art. 

\label{data}

\begin{table}[!htb]
    \small
    \begin{minipage}{.5\linewidth}
      \centering
      \caption{Classification comparison}
        \begin{tabular}{lrrr}
        \toprule
        Paper & Precision & Recall & GSD$^1$\\
        \midrule
        Malof et al. \cite{Malof2017} & 95\% & 80\% & 30 \\
        DeepSolar \cite{Yu2018} & 93.1\% & 88.5\% & 5 \\
        Stamatiou et al. \cite{Stamatiou2018} & ~93\% & ~93\% & - \\
        \bottomrule
        This work & 87.3\% & 87.5\% & 10 \\
        \label{tbl: classificationpaper}
        \end{tabular}
    \end{minipage}%
    \begin{minipage}{.5\linewidth}
      \centering
        \caption{Segmentation comparison}
        \begin{tabular}{lrrr}
        \toprule
        Paper & MAPE$^2$ & mIoU$^3$ & GSD$^1$\\
        \midrule
        Camilo et al. \cite{segsolar} & - & 60\% & 30\\
        DeepSolar \cite{Yu2018} & 24.6\% & - & 5\\
        SolarNet \cite{Hou2020} & - & 90.9\%& 5\\
        \bottomrule
        This work & 18.5\% & 74.1\% & 10\\
        \label{tbl: segpaper}
        \end{tabular}
    \end{minipage} 
    \footnotesize{1) Ground Sampling Distance~(GSD) denotes the native spatial resolution of a dataset and is measured in cm/pixel. 2) Mean Absolute Percentage Error (MAPE). 3) mean Intersection-over-Union (mIoU)}
\end{table}

\textbf{Classification.} 
The classification model achieves a precision of 87.3\% and a recall of 87.5\% on the test set (for an imbalance rate of 7 and a threshold of 0.68). 
Without a benchmark dataset, our classification model appears to perform at a similar level as \cite{Yu2018} but notably worse than \cite{Stamatiou2018}.

\textbf{Segmentation.} The segmentation model for PV panel size estimation yields a Mean Absolute Percentage Error~(MAPE) of 
18.5\% and an Intersection-over-Union~(IoU) of 74.1\% on the test set (for a weight of 0.3 and a threshold of 0.5). In addition, we find that our segmentation estimation is unbiased since it underestimates the overall PV system area in the test set by only 3.9\%. Without a benchmark dataset, our segmentation model appears to perform better than \cite{Yu2018} but worse than \cite{Hou2020}.

It must be noted that each of the related approaches uses different datasets and comparability is therefore limited. Yet, the results indicate that our pipeline's performance is in a similar range as the current state of the art.

\subsection{Improved PV Capacity Estimation with Rooftop Tilt Angles}

To demonstrate the value of enriching the segmented PV systems with 3D building data, we show how information on the rooftop tilt improves the estimation of PV system capacities. Overall, in our study area, the roof surfaces that are equipped with PV systems are tilted between 0 (flat roofs) and 80 degrees. The majority is tilted between 10 and 15 degrees; however, there is considerable variation. A histogram of the rooftop tilts is displayed in Figure \ref{fig:tilt_angles} in the appendix. We now use 120 PV systems that are detected by our pipeline and compare their estimated capacities with the entries in the official registry. \Cref{tbl: tiltimpact} displays the results with and without the incorporation of rooftop tilt angles. We find a MedAPE of 25.9\% when neglecting rooftop tilt. When incorporating it, the error is considerably reduced to 16.1\%. Overall, this demonstrates the rooftop tilt's importance for PV capacity estimation and associated applications, such as PV supply predictions or nowcasting. Furthermore, this analysis underlines the value of enriched PV registries as presented in this paper.

\begin{table}[!htb]
    \small
    \centering
    \begin{minipage}{.5\linewidth}
    \centering
      \caption{PV capacity estimation error}
        \begin{tabular}{lr}
            \toprule
             Approach & MedAPE*\\
            \midrule
            This work (no tilt) & 25.9\% \\
            This work (including tilt) & 16.1\% \\
            \bottomrule
            \multicolumn{2}{l}{\footnotesize{*denotes the Median Absolute Percentage Error}}
        \label{tbl: tiltimpact}
        \end{tabular}
    \end{minipage}%
    \begin{minipage}{.55\linewidth}
        \centering
        \caption{German official registry vs. this work}
        \begin{tabular}{lrr}
        \toprule
        Dataset & \# entries & Capacity [$kWp$]\\
        \midrule
        Official registry & 1,509 & 57,219\\
        Official registry$^1$ & 1,509 & 29,758\\
        This work & 1,211 & 32,386\\
        \bottomrule
        \multicolumn{3}{l}{1) corrected for duplicates and erroneous capacities}
        \label{tbl:registercomp}
        \end{tabular}
    \end{minipage} 
\end{table}

\subsection{Comparison with Germany's Official Registry}
\label{sec:ourapproachVSregister}
\label{results}

In total, we localize 1,211 PV systems in Bottrop, while Germany’s official PV registry lists around 1,509 PV systems. When comparing the officially registered capacity with the estimated installation capacity, the official registry yields 57,219kWp, while we identify 32,386kWp taking the tilt angle into account. Using our automated register to analyze these significant differences, we are able to unveil the following discrepancies:
\begin{itemize}
    \item \textbf{Duplicated entries.} In total, 3.2\% of the official registry’s entries turn out to be duplicates. These duplicates potentially result from PV systems that are reported multiple times by their owners or other errors during data collection. 
    \item \textbf{Erroneous capacities.} By comparing our capacity estimations with the registered capacities, we notice entries with substantially inflated installation capacities.
    \item \textbf{Multiple entries per address.} For some addresses, multiple PV systems are registered. This may arise from the fact that each commissioning of a PV system is registered separately in the official registry. While this is not an error per se, it may still be impractical for analyses based on the registry.
\end{itemize}
After correcting for duplicates and inflated capacities, the official registry yields a total installation capacity of 29,758$kWp$, as summarized in \Cref{tbl:registercomp}. Hence, our pipeline overestimates the total installation capacity by only 8.8\%. By investigating further, we are able to identify additional discrepancies. For this analysis, we consider only PV systems above 30kWp, which corresponds to 160 entries in the official registry:
\begin{itemize}
    \item \textbf{False addresses.} 24 out of 160 entries are listed with a false address in the official registry. This issue arises particularly for larger PV systems, e.g. at industrial facilities, where the address of the system owner is recorded instead of the address of the actual PV system location.
    \item \textbf{Missing entries.} Our pipeline identifies 21 PV systems that are not listed in the official registry.
\end{itemize}
Overall, these results demonstrate that the presented automated PV registry can help to improve PV registries based on self-reported data. However, it must be noted that automated PV registries cannot yet replace self-reported registers: 16 out of the 160 entries above 30kWp were not detected by our pipeline or the capacity estimation failed. 

\section{Discussion and Outlook}
\label{discussion}

Based on our results, we are confident that our approach can provide a good alternative to generate, update, and enhance registries which rely on self-reporting.
Furthermore, by providing detailed information on PV installations, our approach enables improved solar generation modeling and supports PV supply forecasting as well as nowcasting. This facilitates improved integration studies for smart grid components such as electric vehicle charging infrastructure, additional PV capacity, and grid reinforcements.
It should be noted that our approach relies on the availability of up-to-date aerial imagery and 3D building data which might not be readily available. Moreover, policy-relevant information such as institutional ownership of PV systems cannot be directly obtained through our approach but could be collected through other means such as automated web search.





\bibliographystyle{plainnat}
\bibliography{main}

\begin{thebibliography}{17}
\providecommand{\natexlab}[1]{#1}
\providecommand{\url}[1]{\texttt{#1}}
\expandafter\ifx\csname urlstyle\endcsname\relax
  \providecommand{\doi}[1]{doi: #1}\else
  \providecommand{\doi}{doi: \begingroup \urlstyle{rm}\Url}\fi

\bibitem[Biljecki et~al.(2015)Biljecki, Heuvelink, Ledoux, and
  Stoter]{tiltazimuth}
Filip Biljecki, Gerard B~M Heuvelink, Hugo Ledoux, and Jantien Stoter.
\newblock {Propagation of positional error in 3D GIS: estimation of the solar
  irradiation of building roofs}.
\newblock \emph{International Journal of Geographical Information Science},
  29\penalty0 (12):\penalty0 2269--2294, December 2015.
\newblock \doi{10.1080/13658816.2015.1073292}.

\bibitem[Bundesnetzagentur(2019)]{mastr}
Bundesnetzagentur.
\newblock Markstammdatenregister, April 2019.
\newblock URL
  \url{https://www.marktstammdatenregister.de/MaStRHilfe/subpages/fristen.html}.
\newblock last accessed on 09/24/20.

\bibitem[Camilo et~al.(2018)Camilo, Wang, Collins, Bradbury, and
  Malof]{segsolar}
Joseph Camilo, Rui Wang, Leslie~M Collins, Kyle Bradbury, and Jordan~M Malof.
\newblock Application of a semantic segmentation convolutional neural network
  for accurate automatic detection and mapping of solar photovoltaic arrays in
  aerial imagery.
\newblock \emph{arXiv preprint arXiv:1801.04018}, 2018.

\bibitem[{CBS DeepSolaris}(2020)]{CBS2020}
{CBS DeepSolaris}.
\newblock {Automatically detect solar panels with aerial photos}, 2020.
\newblock URL
  \url{https://www.cbs.nl/en-gb/about-us/innovation/project/automatically-detect-solar-panels-with-aerial-photos}.
\newblock last access on 2020-07-10.

\bibitem[Chen et~al.(2017)Chen, Papandreou, Schroff, and
  Adam]{chen2017rethinking}
Liang-Chieh Chen, George Papandreou, Florian Schroff, and Hartwig Adam.
\newblock Rethinking atrous convolution for semantic image segmentation.
\newblock \emph{arXiv preprint arXiv:1706.05587}, 2017.

\bibitem[de~Hoog et~al.(2020)de~Hoog, Maetschke, Ilfrich, and
  Kolluri]{deHoog2020}
Julian de~Hoog, Stefan Maetschke, Peter Ilfrich, and Ramachandra~Rao Kolluri.
\newblock Using satellite and aerial imagery for identification of solar pv:
  State of the art and research opportunities.
\newblock In \emph{Proceedings of the Eleventh ACM International Conference on
  Future Energy Systems}, page 308–313. Association for Computing Machinery,
  2020.
\newblock ISBN 9781450380096.

\bibitem[Energieexperten(2020)]{sizepv}
Energieexperten.
\newblock Pv modul-größen im Überblick, 2020.
\newblock URL
  \url{https://www.energie-experten.org/erneuerbare-energien/photovoltaik/solarmodule/groesse.html}.
\newblock last accessed on 09/24/20.

\bibitem[Hou et~al.(2020)Hou, Wang, Hu, Yin, Huang, and Wu]{Hou2020}
Xin Hou, Biao Wang, Wanqi Hu, Lei Yin, Anbu Huang, and Haishan Wu.
\newblock {SolarNet: A Deep Learning Framework to Map Solar PowerPlants In
  China From Satellite Imagery}.
\newblock 2020.
\newblock URL \url{https://arxiv.org/pdf/1912.03685.pdf}.

\bibitem[Kolbe et~al.(2005)Kolbe, Gr{\"o}ger, and Pl{\"u}mer]{citygml}
Thomas~H Kolbe, Gerhard Gr{\"o}ger, and Lutz Pl{\"u}mer.
\newblock Citygml: Interoperable access to 3d city models.
\newblock In \emph{Geo-information for disaster management}, pages 883--899.
  Springer, 2005.

\bibitem[Laboratory()]{openpv}
National Renewable~Energy Laboratory.
\newblock The open pv project.

\bibitem[{Malof} et~al.(2017){Malof}, {Collins}, and {Bradbury}]{Malof2017}
J.~M. {Malof}, L.~M. {Collins}, and K.~{Bradbury}.
\newblock A deep convolutional neural network, with pre-training, for solar
  photovoltaic array detection in aerial imagery.
\newblock In \emph{2017 IEEE International Geoscience and Remote Sensing
  Symposium (IGARSS)}, pages 874--877, 2017.

\bibitem[Malof et~al.(2017)Malof, Collins, and Bradbury]{cnnsolar1}
Jordan~M Malof, Leslie~M Collins, and Kyle Bradbury.
\newblock A deep convolutional neural network, with pre-training, for solar
  photovoltaic array detection in aerial imagery.
\newblock In \emph{2017 IEEE International Geoscience and Remote Sensing
  Symposium (IGARSS)}, pages 874--877. IEEE, 2017.

\bibitem[Mayer et~al.(2020)Mayer, Wang, Arlt, Rajagopal, and
  Neumann]{Mayer2020}
Kevin Mayer, Zhecheng Wang, Marie-Louise Arlt, Ram Rajagopal, and Dirk Neumann.
\newblock {DeepSolar for Germany : A deep learning framework for PV system
  mapping from aerial imagery}.
\newblock In \emph{IEEE Smart Energy Systems and Technologies, Istanbul,
  Turkey, September 07 - 09, 2020}. IEEE, 2020.

\bibitem[Nordrhein-Westfalen(2020{\natexlab{a}})]{3ddaten}
Land Nordrhein-Westfalen.
\newblock 3d-gebäudemodelle, May 2020{\natexlab{a}}.
\newblock URL
  \url{https://www.bezreg-koeln.nrw.de/brk_internet/geobasis/3d_gebaeudemodelle/index.html}.
\newblock last accessed on 05/24/20.

\bibitem[Nordrhein-Westfalen(2020{\natexlab{b}})]{aerialimagerynrw}
Land Nordrhein-Westfalen.
\newblock Digitale orthophotos, May 2020{\natexlab{b}}.
\newblock URL
  \url{https://www.bezreg-koeln.nrw.de/brk_internet/geobasis/luftbildinformationen/aktuell/digitale_orthophotos/index.html}.
\newblock last accessed on 09/24/20.

\bibitem[Stamatiou(2018)]{Stamatiou2018}
Kostas Stamatiou.
\newblock {How We Used Deep Learning to Identify Solar Panels on 15 Million
  Buildings}, 2018.
\newblock last access on 2020-07-10, [Online]. Available:
  https://blog.maxar.com/earth-intelligence/2018/how-we-used-deep-learning-to-identify-solar-panels-on-15-million-buildings.

\bibitem[Yu et~al.(2018)Yu, Wang, Majumdar, and Rajagopal]{Yu2018}
Jiafan Yu, Zhecheng Wang, Arun Majumdar, and Ram Rajagopal.
\newblock {DeepSolar: A Machine Learning Framework to Efficiently Construct a
  Solar Deployment Database in the United States}.
\newblock \emph{Joule}, 2\penalty0 (12):\penalty0 2605--2617, 2018.

\end{thebibliography}

\section{Appendices}

\subsection{Training DeepSolar GER for PV system detection}
\label{sec: traindeepsolar}

We use the pre-trained DeepSolar network for PV system detection in Germany and train it for 25 epochs as described in \cite{Mayer2020}. We used a batch size of 8, an imbalance rate of 7, and an initial learning rate of 0.0001. Training stops as soon as the binary cross-entropy does not decrease for 5 subsequent epochs.

\subsection{Training DeepLabv3 for PV system segmentation}
\label{sec: traindeeplab}

Due to class imbalance, we use a weight factor to emphasize PV panel pixels in the binary cross-entropy calculation. The result of the segmentation model is a $320 \times 320$ probability density map indicating the likelihood that a certain pixel in the image depicts a PV system. The model is trained for 25 epochs with a batch size of 10, however, training stops as soon as the binary cross-entropy does not decrease for 10 subsequent epochs. To find the estimated area covered by PV systems in an image, we normalize the resulting segmentation map so all values lie in $[0,1]$ and search for the threshold at which the model yields the lowest MAPE on the validation set. The obtained threshold is used for the test set and for inference for all images in North Rhine-Westphalia. \\

\subsection{Dataset statistics}
\label{dataset_statistics}

\begin{table}[htb]
  \caption{Overview of classification dataset}
  \label{tab:classification_dataset}
  \centering
  \begin{tabular}{lllll}
    \toprule
    \cmidrule(r){1-2}
    Split & Positive & Negative & Total & Share [\%] \\
    \midrule
    Training & 1,814 & 36,790 & 38,604 & 35.8  \\
    Validation & 216 & 23,929 & 24,145 & 22.4    \\
    Test & 385 & 44,675 & 45,060 &  41.8 \\
    \midrule
    Total & 2,415 & 105,394 & 107,809 & 100 \\
    \bottomrule
  \end{tabular}
\end{table}

\begin{table}[htb]
  \caption{Overview of segmentation dataset}
  \label{tab:segmentation_dataset}
  \centering
  \begin{tabular}{lll}
    \toprule
    \cmidrule(r){1-2}
    Split & Positive & Share [\%] \\
    \midrule
    Training & 3,222  & 80\\
    Validation & 403 & 10\\
    Test & 403 & 10\\
    \midrule
    Total & 4,028 & 100\\
    \bottomrule
  \end{tabular}
\end{table}
\newpage
\subsection{Tilt Histogram: City of Bottrop}

\label{sec: tiltdist}
\begin{figure}[!htb]
  \begin{center}
    \includegraphics[ scale=0.5]{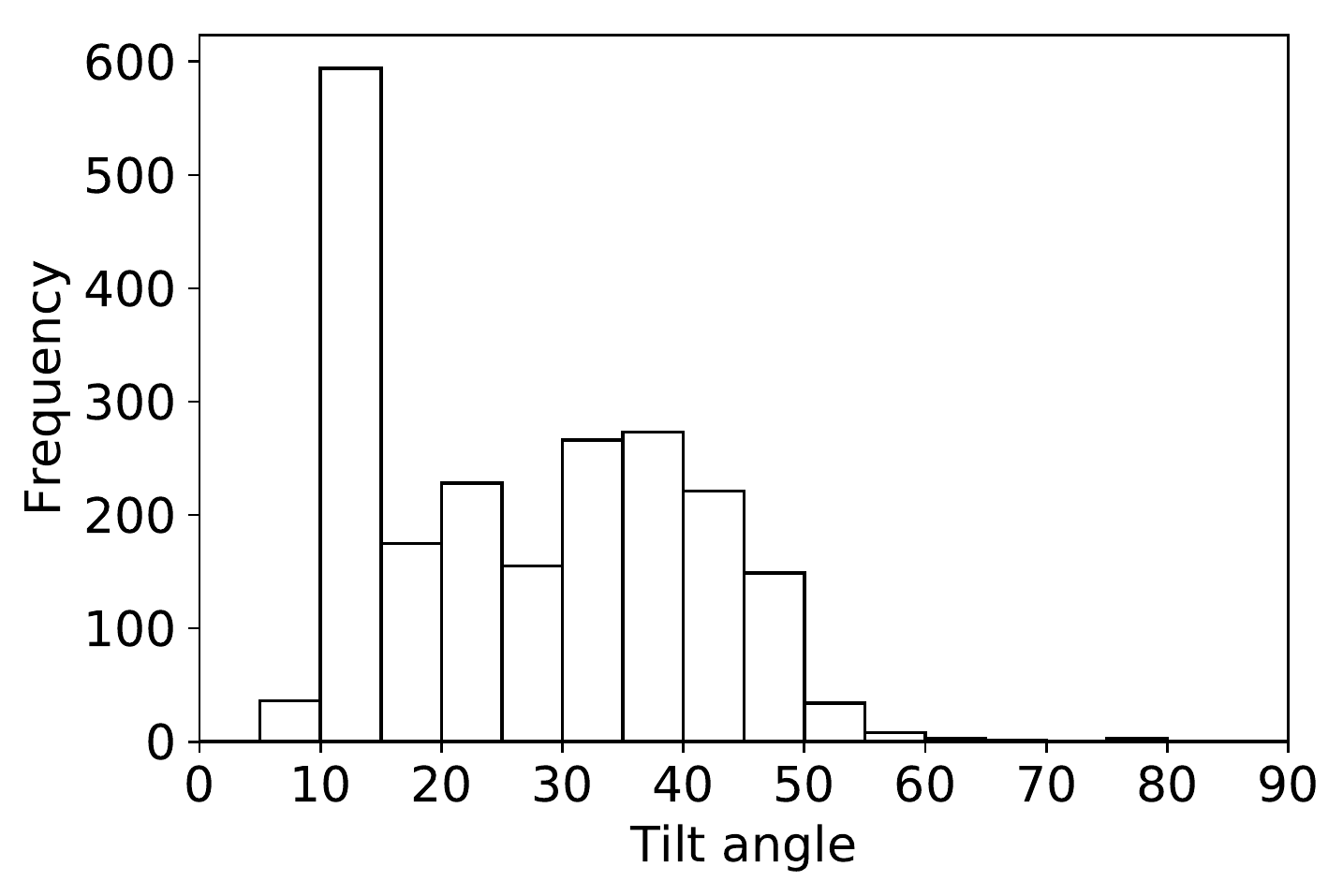}
  \end{center}
  \caption{Histogram of tilt angles in the city of Bottrop (non-flat rooftops). 69.7\% of PV instances are installed on flat rooftops. Each PV instance represents a single segmented PV polygon. }
  \label{fig:tilt_angles}
\end{figure}

\end{document}